# A Spectral-Spatial Dependent Global Learning Framework for Insufficient and Imbalanced Hyperspectral Image Classification

Qiqi Zhu, Weihuan Deng, Zhuo Zheng, Yanfei Zhong*, *Senior Member, IEEE*, Qingfeng Guan, Weihua Lin, Liangpei Zhang, *Fellow, IEEE*, Deren Li, *Senior Member, IEEE*

*Abstract*—Deep learning techniques have been widely applied to hyperspectral image (HSI) classification and have achieved great success. However, the deep neural network model has a large parameter space and requires a large number of labeled data. Deep learning methods for HSI classification usually follow a patchwise learning framework. Recently, a fast patch-free global learning (FPGA) architecture was proposed for HSI classification according to global spatial context information. However, FPGA has difficulty extracting the most discriminative features when the sample data is imbalanced. In this paper, a spectral-spatial dependent global learning (SSDGL) framework based on global convolutional long short-term memory (GCL) and global joint attention mechanism (GJAM) is proposed for insufficient and imbalanced HSI classification. In SSDGL, the hierarchically balanced (H-B) sampling strategy and the weighted softmax loss are proposed to address the imbalanced sample problem. To effectively distinguish similar spectral characteristics of land cover types, the GCL module is introduced to extract the long short-term dependency of spectral features. To learn the most discriminative feature representations, the GJAM module is proposed to extract attention areas. The experimental results obtained with three public HSI datasets show that the SSDGL has powerful performance in insufficient and imbalanced sample problems and is superior to other state-of-the-art methods. Code can be obtained at: https://github.com/dengweihuan/SSDGL.

*Index Terms*—deep learning, patchwise, hyperspectral image classification, imbalanced sample, feature representations.

## I. INTRODUCTION

With the development of remote sensing techniques, numerous hyperspectral images (HSIs) can be obtained with abundant spectral information. HSIs are composed of narrow and contiguous spectral bands in the electromagnetic spectrum [1]–[3]. Because of the richness of spectral information, it has a wide range of applications in various fields, such as land-cover detection, agricultural development, environmental protection and urban planning [4]. Hyperspectral image classification, which aims to assign a unique label to each pixel, plays an essential role in the interpretation of hyperspectral remote sensing images [5, 6]. However, due to the high dimensionality of HSI and limited labeled data, data redundancy and the Hughes phenomenon often arise and pose a major challenge for HSI classification. To further improve the HSI classification performance, many studies have been undertaken over the years [7]–[9]. Traditional hyperspectral image classification usually carries out feature extraction first and then classifies HSI by various classifiers, such as multinomial logistic regression (MLR), maximum likelihood classification (MLC), and support vector machine (SVM) [10, 11]. Numerous studies have shown that introducing spatial information into the classification process can effectively improve the performance of HSI classification. Spectral-spatial based methods, such as simple linear iterative clustering (SLIC), extended morphological profiles (EMP) and the Gabor filter, have been proposed to extract both the spectral and spatial features of HSIs [12, 13]. However, these spectral-spatial features and hyperparameters are selected based on prior information, and the classification performance is limited by the number of training samples.

With the development of deep learning, convolutional neural network (CNN)-based methods have attracted great attention for hyperspectral image classification [14]–[16]. As a data-driven automatic feature learning framework, it can achieve end-to-end training and automatically extract the spectral-spatial features of the images. Many supervised classification methods based on deep learning are used to extract the spectral and spatial features, such as 3D-CNN, multiscale convolutional neural networks [17, 18]. Recurrent neural networks (RNNs) and convolutional recurrent neural networks (CRNNs), which can learn the long short-term spectral dependencies, are widely used in hyperspectral image classification [19, 20]. In addition, residual networks, capsule networks, double-branch networks, and other novel networks have been widely applied in HSI classification and have achieved great classification accuracy with sufficient labeled samples [21]. However, these methods only consider the labeled samples and ignore the spectral-spatial information of unlabeled samples. To make full use of the unlabeled data, many semisupervised classification methods have been proposed. It can be observed that HSIs have strong spatial

Manuscript received September 18, 2020; revised February 15, 2021; accepted March 23, 2021. This work was supported by National Natural Science Foundation of China under Grant No. 41901306, and a Grant from State Key Laboratory of Resources and Environmental Information System. (Corresponding author: Yanfei Zhong).

The authors are with the School of Geography and Information Engineering, China University of Geosciences, Wuhan 430079, China, and also with the State Key Laboratory of Information Engineering in Surveying, Mapping and Remote Sensing, Wuhan University, Wuhan 430079, China, (e-mail: zhuqq@cug.edu.cn, dengweihuan@cug.edu.cn, guanqf@cug.edu.cn, zhongyanfei@whu.edu.cn, zlp62@whu.edu.cn, drli@whu.edu.cn ) homogeneity, which means the adjacent pixels are likely the same class [22]. Therefore, the superpixel-based methods can effectively extract the deep spatial information for HSI classification. Moreover, the spatial size and shape of the superpixels are adaptive [23]. To further solve the problem of the small samples, the generative adversarial network was proposed to generate pseudolabeled samples and make the distribution of the fake samples closer to real data distribution [24]–[26]. However, most existing methods are patch-based learning frameworks, and the neighbor region of each pixel in HSIs is considered as the input data of the network. The high computational complexity is unavoidable since there are large overlapped areas among adjacent patches.

To solve the above problems, a fast patch-free global learning (FPGA) framework was proposed to maximize the exploitation of the global spatial information according to the long-range spatial dependency [27]. FPGA has achieved great classification accuracy on public datasets and reduced the redundant calculations. However, the number of training samples per class is the same, and there are completely different labeled samples in each hierarchical training dataset, which is not suitable for datasets with insufficient and imbalanced samples, such as the Indian Pines dataset. Moreover, the categories with few samples have a small amount of weight in the loss calculation, and it is difficult for FPGA to classify these categories well. Since some land-cover types are difficult to distinguish by visual interpretation, the long-tail distribution issue of hyperspectral image datasets has arisen and seriously limits the classification performance.

In this paper, to extract the deep spectral-spatial features and solve the sample problem of insufficiency and imbalance, a spectral-spatial dependent global learning (SSDGL) framework combining global convolutional long short-term memory (GCL) and global joint attention mechanism (GJAM) is proposed. Compared with the FPGA sampling strategy, a hierarchically balanced (H-B) sampling strategy is proposed to obtain enough hierarchical data and balance minibatch per class. The weighted softmax with cross entropy loss is used to give each class an equal probability of being selected. The novel sampling strategy and loss strategy can effectively solve the class imbalance problem. The baseline of the proposed framework is an encoder-decoder architecture (SegNet) [28], which has achieved good classification performance in image segmentation. Furthermore, the GCL module is introduced to extract the long short-term spectral dependent features and obtain the interrelation among the local pixels. The GJAM module is utilized to extract more discriminative feature representations.

The main contributions of this paper are as follows.
1) A spectral-spatial dependent global learning (SSDGL) framework is proposed for HSI classification. To solve the insufficient and imbalanced sample problems, a hierarchically balanced sampling strategy is utilized to generate stochastic hierarchical training sample data. The proposed sampling strategy reduces the overall training times and speeds up model convergence. The weighted softmax with cross entropy loss is introduced to reduce the weight of easy-to-classify samples so that the model focuses more on hard-to-classify samples during training. All pixels are used for the convolution operation at the same time, which solves the problem of the limited patch size.
2) To extract the detailed spectral-spatial information of the whole image, GCL is proposed to capture the long short-term spectral dependent features and leverage convolutional kernel to extract interrelations among the local pixels. GCL is a sequence-to-sequence learning method, and the gated recurrent units are utilized to extract deep spectral and spatial features. This module can effectively distinguish similar land covers by extracting the intrinsic spectral-spatial dependency.
3) To further extract the most discriminative feature representation, a global joint attention mechanism is designed to reweight and model the extracted features. This module is composed of a spectral attention mechanism and a spatial attention mechanism. The spectral attention mechanism can selectively emphasize informative spectral features and suppress less-useful ones. The spatial attention mechanism is introduced to extract the short-term spatially dependent features and emphasize the key regions.

The rest of this paper is organized as follows. Section II discusses the related work. Section III provides a detailed description of the SSDGL framework for insufficient and imbalanced HSI classification. A description of the datasets and an analysis of the experimental results are presented in Section IV. The sensitive parameters are discussed in Section V. Finally, the conclusions are drawn in Section VI.

## II. BACKGROUND

In recent years, deep learning techniques have achieved great success in the field of remote sensing. The CNN-based classification methods and fully convolutional network (FCN)-based classification methods have been applied to HSI classification successfully. Moreover, insufficient and imbalanced sample problems have become a research hot spot in image classification.

### A. CNN-based classification

CNN-based classifications are regarded as common feature learning methods, which have a significant advantage in accuracy, and classifications are performed in an end-to-end manner [29]. To facilitate feature extraction and train the classifiers, HSI pixel patches are first generated from the original image by a sliding window with a fixed size. The spectral-spatial residual network (SSRN) and double branch multi-attention mechanism (DBMA) network were proposed to extract the deep spectral and spatial features. As the layer goes deeper, the features in the model become more abstract and more robust. The extracted spectral-spatial features were flattened into vectors and fused for classification [30, 31]. Mou et al. proposed RNNs extracting the spectral dependency among adjacent wavebands and regarded the HSI as sequential data [16]. Subsequently, CRNN was proposed to highlight long-term dependency between nonadjacent channel features. To consider both spatial and spectral information, the ConvLSTM was proposed to extract dependent features of the spectrums and geometric [32, 33].

Although these patchwise methods have achieved significant classification accuracies, redundant computation on the overlapping areas between adjacent patches is inevitable. The main bottleneck is that the traditional convolutional neural networks first divide HSI into patches and classify each patch into one corresponding label rather than directly classifying the whole image [34, 35].

### B. FCN-based classification

Fully convolutional neural networks were first proposed for the task of image segmentation and achieved great success. The fully connected layers were replaced by convolutional layers with a kernel size of $1 \times 1$. U-Net is the most representative fully convolutional network. It consists of a contracting path to capture context and a symmetric expanding part to precise localization [36]. SegNet is a network based on FCN and consists of an encoder network, a corresponding decoder network followed by a pixelwise classification layer. The feature pyramid network (FPN), which improved on U-Net, was proposed to capture multiscale information. It composed of a bottom-up pathway, a top-down pathway and lateral connections [37]. Inspired by the FPN model, the DeepLab networks combining atrous convolution, spatial pyramid pooling, and fully connected CRFs have been proposed and used to extract feature representations with different scales [38].

The purpose of hyperspectral image classification is the same as semantic segmentation tasks, which is to assign a unique label to each pixel. However, it is not feasible to directly transfer the semantic segmentation networks to HSI classification tasks, since the training sample of HSI datasets is highly sparse and only contains a set of discrete labeled pixels rather than a group of labeled images [39]. Xu et al. [35] proposed a spectral-spatial fully convolutional network (SSFCN) to perform feature extraction and semantic segmentation in an end-to-end manner. A novel mask matrix was proposed to deal with the high sparse training samples in HSIs. To solve the problem of model convergence and mine the global spatial context information, Zheng et al. [27] proposed a fast patch-free global learning framework (FPGA) for HSI classification. FPGA is a deep convolutional encoder-decoder architecture, the global stochastic stratified (GS²) sampling strategy was introduced to obtain diverse gradients to guarantee the convergence of the FCN in the FPGA framework. The FreeNet model in FPGA was proposed to avoid redundant computation on the overlapping areas between patches. The comparison of the patch-based and patch-free methods is shown in Fig. 1.

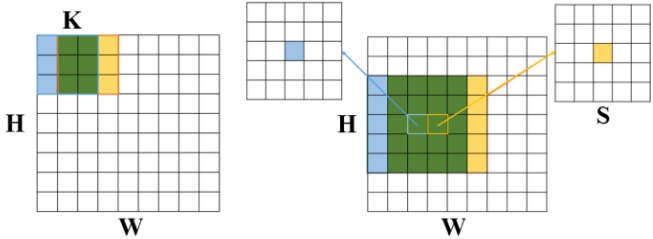

Fig. 1. Convolution operation of patch-based and patch-free.

The green area is the overlapping area. $H$ and $W$ are the spatial sizes of the input data, $K$ is the convolutional kernel size, and $S$ is the pixel patch size. It can be seen that the patchwise classification methods need to generate a patch for each pixel, which will generate redundant calculations in the model inference. However, the input data of global learning methods is the whole image, and there are no pixel patches or center pixels, which speeds up the model inference. Since the convolution operation considers all pixels of the whole image at the same time, the classification performance has been greatly improved [40].

### C. Insufficient and imbalanced sample problems

It is well known that the problem of limited training samples is one of the major obstacles that affect the accuracy of HSI classification. Since transfer learning methods using the pretrained deep learning network from the relevant domain to aid learning, few samples are required to fine-tune the model [41]. Yang et al. pretrained a two-channel CNN network on source HSI, which contains sufficient labeled samples. Then the bottom layers of the pretrained network were transferred to the target network as initialization, and the top layers were randomly initialized [42]. Pan et al. [9] proposed a small-scale data-based method, the multi-grained network (MugNet), which can obtain the fined spectral and spatial features. It used all the unlabeled samples to learn convolution kernels and build a lightweight network that does not include many hyperparameters for tuning. Fang et al [6] designed a lightweight 3D convolutional neural network and a novel clustering strategy for learning the deep discriminative feature and perform semi-supervised classification. Mei et al. [43] proposed a 3 dimensional (3D) convolutional autoencoder (3D-CAE) to maximally obtain the most discriminative spectral and spatial information for feature extraction. This framework can learn the deep features in an unsupervised mode and network trained without labeled training samples.

Moreover, the problem of long-tail distribution poses great challenges for deep learning, and it has attracted increasing attention in computer vision. Existing solutions usually have a difference in sampling strategy and classifiers [44]. For most sampling strategies presented below, the probability $p_j$ of sampling a data point from class $j$ is given by:

$$p_j = \frac{n_j^q}{\sum_{i=1}^{C} n_i^q} \qquad (1)$$

where $q \in [0,1]$ and $C$ is the number of training classes.

The instance-balanced sampling, class-balanced sampling, and square-root sampling all achieved good results for long-tailed recognition, where $q$ is set to 1, 0, and 1/2 respectively. Most of the methods trained the classifiers to rectify the decision boundaries on the head- and tail-classes via fine-tuning, and optimize parameters by loss reweighting strategies [45, 46]. Kang et al. [47] proposed a decoupled representation learning and classification strategy to compare the performance differences of different sampling strategies and classifiers for classification.

Several important questions need to be considered. How can the global learning framework consider the relationship between long short-range bands? In addition, long-tail distribution exists in the HSI datasets and limits the performance of HSI classification, so how can we address insufficient and imbalanced training data problems? To overcome the aforementioned issues, we propose the SSDGL framework for HSI classification using small training samples.

## III. SSDGL: SPECTRAL-SPATIAL DEPENDENT GLOBAL LEARNING FRAMEWORK

To extract the spectral relationship among different bands and the spatial correlation of all pixels, the SSDGL framework is proposed for hyperspectral image classification. This is an ensemble learning method that combines spectral, structural, and semantic features. The most discriminative feature representations are learned by the global convolutional long short-term memory integrated with the global joint attention mechanism (GCLAM). The hierarchically balanced sampling strategy is proposed to divide the training data into a hierarchical sequence of training samples, and the weighted softmax with cross entropy loss is introduced to reweight each class probability. The novel sampling strategy and loss function effectively solve sample imbalanced and insufficient problems. The skip connections are utilized to fuse the spatial features from the encoder and the semantic features in the same stage decoder. The overall architecture of the SSDGL is shown in Fig. 2.

### A. Hierarchically balanced sampling strategy

The hierarchically balanced (H-B) sampling strategy was proposed to obtain diverse stochastic gradients, combined with weight decay and learning rate decay to speed up model convergence. In this work, the training sample is the whole image rather than the patches of the local area, and the set of discretely labeled pixels is assigned to a hierarchical sequence to improve the robustness of the model and reduce the training time of SSDGL-Net. Because the labeled pixels of the HSI dataset are insufficient and imbalanced, the labeled samples are divided in a certain ratio and the indices of each category are stored in different lists. The selected sample data are viewed as training data and other sample data as test data. The input data of the SSDGL-Net is a stochastic hierarchical training sample sequence, and the hierarchical training data balanced the number of samples per class. The hyperparameter $\alpha$ represents the number of hierarchical training data, and the number of labeled samples in hierarchical training dataset will affect the speed of model convergence. Each hierarchical training dataset contains all categories, and the mini batch per class is determined by the parameter $\beta$. Within a certain range, the smaller the value of $\beta$ is, the more random gradients can be obtained and the less training time is required. To address the imbalanced sample problem, the weighted softmax loss is introduced to balance the probability of the ground-truth class and focus more on misclassified samples. The category weighting factor $\left( q_j / \sum_{i=0}^{M} q_i \right) * M$ is added to the standard cross-entropy criterion to reduce the relative loss for well-classified samples. $p_j$ is used to balance the probability that selecting a sample data belong to the class $j$.

The number of labeled samples per class in hyperspectral images is different because it is difficult to identify many land-covers by visual interpretation. If using the traditional sampling strategy, the average accuracy and the overall accuracy are greatly limited. Therefore, it is necessary to introduce a novel sampling strategy and a suitable loss function to solve the problem of class long-tail distribution in hyperspectral image datasets. The pseudocode of SSDGL is shown in Algorithm 1.

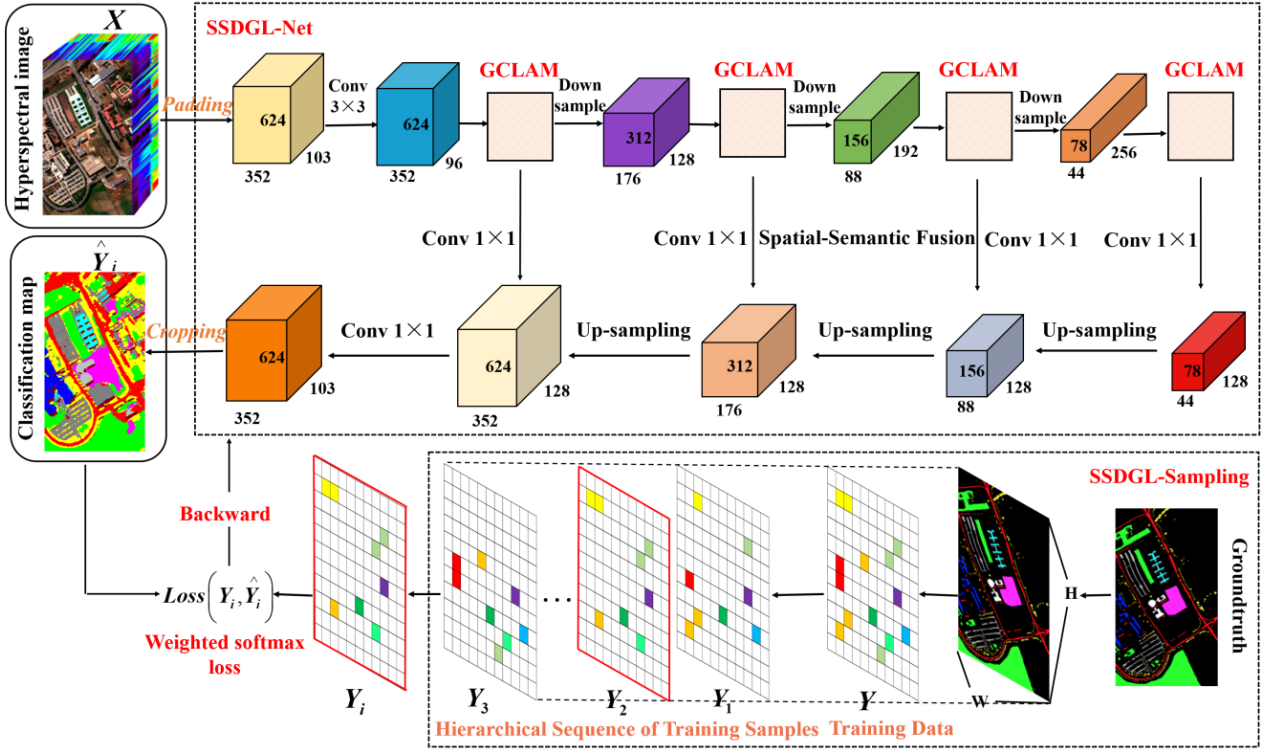

Fig. 2. Flowchart of insufficient and imbalanced HSI classification based on the SSDGL framework.

**Algorithm 1** The pseudocode of SSDGL

**Input:** $N = \{n_i\}_{i=0}^M$ : A set of discrete labeled pixels

   $n_i$ : A set of labeled pixels per class

   $\lambda$ : the training sample ratio

   $M$ : the number of classes

   $\alpha$ : the number of stratified training data

   $\beta$ : mini-batch per class

   $\delta$ : compression factor

   $Mask_{ij}$ : mask matrix

   $y^n$ : predicted scores

   $\hat{y}_{ij}$ : predicted label

**Output:** The parameters of the whole network, $\theta = \{\theta_E, \theta_D\}$,

   classification map $O_{I \times W \times H}$

$R[i, j] \leftarrow [[]]$ an empty matrix

$I_{ij} \leftarrow \{\}$ an empty dict

$w \leftarrow []$ an empty list

$m_i = 1 - \delta^{n_i}$ , $q_i = (1 - \delta) / m_i$

**for** $j = 0$ to $M$ **do**

   $p_j = \dfrac{q_j}{\sum_{i=0}^M q_i} * M$

   $w.push(p_j)$

**end**

**for** $s = 0$ to $\alpha$ **do**

   **for** $k = 0$ to $M$ **do**

      $I_{ij} \leftarrow shuffle(\{n_k \times \lambda\})$

      **while** $len(I_{ij}) < \beta$ **do**

         fetch all samples from $I_{ij}$, $R[i, j] \leftarrow I_{ij}$

         $Mask_{ij} = \begin{cases} 0, & if \ R[i, j] = 0 \\ 1, & if \ R[i, j] > 0 \end{cases}$

      **continue**

         fetch $\beta$ samples from $I_{ij}$, $R[i, j] \leftarrow I_{ij}.pop(\beta)$

         $Mask_{ij} = \begin{cases} 0, & if \ R[i, j] = 0 \\ 1, & if \ R[i, j] > 0 \end{cases}$

   **end**

**end**

**while** training loss has not converged **do**

   $L_w = -\dfrac{1}{\beta \cdot M} \sum \log\left(\dfrac{e_{ij}^{y^n}}{\sum_n e_{ij}^{y^n}} \times w[m]\right) \bullet Mask_{ij}$

   update $\theta_E$ and $\theta_D$ by minimizing $L_w$ through stochastic gradient descent and weight decay:

   $\theta_E \leftarrow \left(1 - \dfrac{\eta\lambda}{n}\right)\theta_E - \eta \dfrac{\partial L_w}{\partial \theta_E}$

   $\theta_D \leftarrow \left(1 - \dfrac{\eta\lambda}{n}\right)\theta_D - \eta \dfrac{\partial L_w}{\partial \theta_D}$

**end while**

$O_{I \times W \times H} \leftarrow \left\{\hat{y}_{ij} \mid 0 \leq i < H, 0 \leq j < W\right\}$

### B. GCLAM

The GCLAM is composed of global convolutional long short-term memory (GCL) and global joint attention mechanisms (GJAM). Since the fully connected LSTM cannot extract local spatial information, it is replaced by ConvLSTM. However, the global spatial context information is ignored with the conventional ConvLSTM, and the importance of extracted features is difficult to emphasize. Hence, GCLAM is proposed to extract the dependency of spectral-spatial features. GCLAM is the most important part of the SSDGL framework, which can extract abundant spectral-spatial features and keep the spatial size of input data unchanged, as shown below.

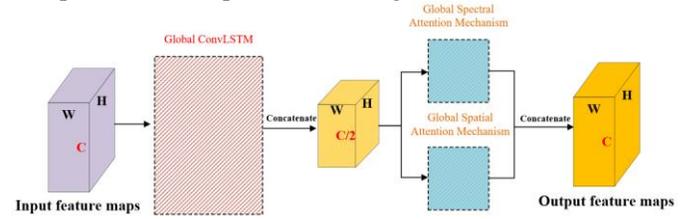

Fig. 3. The architecture of the GCLAM.

a) The global convolutional long short-term memory (GCL) is utilized to extract the spectral dependency according to the global spatial context information. The flowchart is shown in Fig. 4, and the input data is the whole hyperspectral image. The channels of the input data are equally distributed to the $n$ group, where $n$ represents the time step of the GCL. In this work, GCL has two hidden layers. The first hidden layer is utilized to extract the interdependency between long-range features. For example, some different types of crops have similar spectral curves, but the correlation between their green band and the near-infrared band is quite different. Therefore, the spectral dependency of long-range bands can effectively distinguish similar land cover types. The second hidden layer is used to enhance the dependency of adjacent channels. GCL is composed of ConvLSTMCells, as shown in Fig. 5.

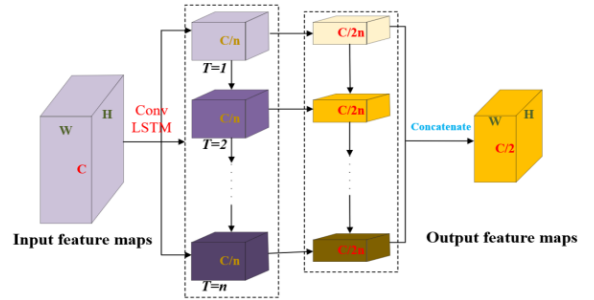

Fig. 4. The architecture of the GCL.

There are four main parts in ConvLSTMCells. a) The forgetting phase selectively forgets information passed from the previous unit and keeps important information. b) The memorizing phase selectively memorizes the input information. c) The output phase determines what information will be output. d) In the convolutional phase, gated units in ConvLSTMCells are equivalent to convolution layers, and it can nonlinearly transform the input features into more discriminative spectral-spatial features.

b) The global joint attention mechanism (GJAM) is composed of a global spectral attention mechanism and global spatial attention mechanism, and they are utilized to estimate the importance of the extracted features. The spectral attention mechanism is used to reweight the spectral features generated by the GCL module and focus on the most discriminative features. The global spatial attention mechanism is utilized to focus on the important local areas and take full advantage of the global spatial context information.

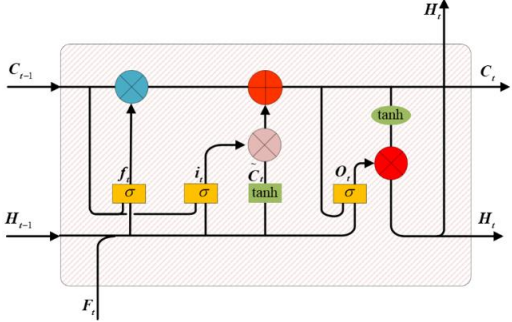

Fig. 5. The architecture of the ConvLSTMCells.

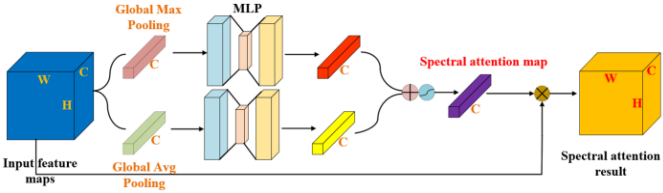

Fig. 6. The flowchart of the global spectral attention mechanism.

1) The global spectral attention mechanism can be regarded as a feature detector, which assigns different weights to each channel. The larger weight is assigned to the meaningful channels and the weight of the meaningless channels is smaller. The global spectral attention mechanism is shown in Fig. 6.

The input feature is the whole image rather than the patches of the local area. The different spectral features are obtained through the global maximum pooling layer and the global average pooling layer. They are represented as $F_{avg}^c$ and $F_{max}^c$. The input image is passed through the pooling layers, and two spectral vectors are generated. The output features are fed to a three-layer perceptron and two output feature vectors are generated. Then, the output feature vectors are merged using an elementwise summation operation. Finally, the output feature vector is multiplied with the input image. It can be expressed by the following formula:

$$M_c(F) = \sigma(MLP(AvgPool(F)) + MLP(MaxPool(F)))$$
$$= \sigma(W_1(W_0(F_{avg}^c)) + W_1(W_0(F_{max}^c))) \quad (2)$$

2) In the HSI data cube, the adjacent pixels are likely to form an area and they belong to the same class. The spatial attention mechanism focuses on attention areas and reweights the generated attention areas, as shown in Fig. 7. The global spatial attention also used the maximum pooling layer and average pooling layer to extract different spatial information. The input features are fed to the pooling layers and two spatial feature maps are generated. Connect two feature maps through concatenation operation, and generate a one-dimensional feature map through the activation function and convolution operation. Multiply the one-dimensional feature map with the input feature maps channel by channel. The spatial features are reweighted and the spatial size is unchanged. It is expressed by the following formula:

$$M_s(F) = \sigma(f^{N \times N}(AvgPool(F); MaxPool(F)))$$
$$= \sigma(f^{N \times N}(F_{avg}^c; F_{max}^c)) \quad (3)$$

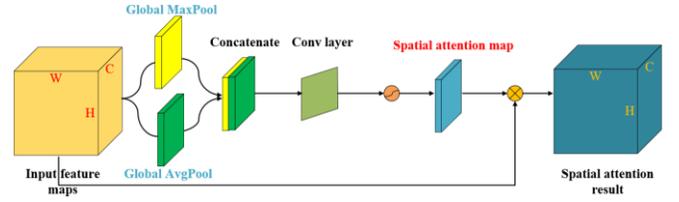

Fig. 7. The flowchart of the global spatial attention mechanism.

Generally, the data with same features will gather together to form an area. To highlight these areas, the spatial attention mechanism is necessary. Since some land covers have large spatial variability, the global learning method is necessary.

## IV. EXPERIMENTS RESULTS AND ANALYSIS

To quantitatively and qualitatively analyze the classification performance of the proposed models, it was compared with some state-of-the-art methods for HSI classification, which include support vector machines with radial basis function kernel (RBF-SVM), semisupervised convolutional neural network (SS-CNN) [48], spectral-spatial residual network (SSRN) [30], double-branch multiattention mechanism network (DBMA) [31], MCNN- CONVLSTM [49], U-Net [36] and FPGA [27]. Extensive experiments were conducted on three datasets: the 16-class Indian Pines dataset, the 9-class Pavia University dataset, and the 15-class Houston University dataset. These datasets are utilized to validate the effectiveness of the proposed method in the cases of imbalanced sample data, high spatial resolution data, and a small number of sample data. All experiments were carried out based on the PyTorch library on a GeForce RTX 2080 Ti graphics card.

### A. Experimental Settings

1) Model parameters: This is a framework based on encoding and decoding. To make the size of the input image meet the downsampling requirements, the input image was increased to a multiple of 16 and padded with zero. The group number of group normalization was set to 4, so the output channels of each layer must be a multiple of 4. Since this framework employs skip connections between the spatial features in the encoder and semantic features in the decoder, the channel of skip connections was set to 128 for feature fusion.

2) Optimized parameters: The time step of the global convolutional long short-term memory was set to 8, and the size of the convolutional kernel was set to 5. The optimizer plays an important role in the training processes of the deep

CNN model and affected the model convergence [50]. The proposed framework was trained in 600 epochs training processes for each dataset and using gradient descent with momentum, where the initial learning rate was set to 0.005 and multiplied by $\left(1 - \dfrac{iter}{max\_iter}\right)^{power}$ with $power = 0.8$ and $max\_iter = 1{,}000$. The momentum was set to 0.9 and the weight decay rate was set to 0.001.

3) Metrics: To evaluate the performance of the proposed methods, four commonly used quantitative metrics were adopted: the accuracy of each class, the overall accuracy (OA), the average accuracy (AA), and the kappa coefficient (Kappa). To eliminate the deviation introduced by randomly choosing training samples, each experiment was run ten times, and the mean values of each evaluation criterion are reported.

### B. Experiment 1: Indian Pines Dataset

The Indian Pines dataset was acquired in 1992 by the Airborne Visible/Infrared Imaging Spectrometer (AVIRIS) sensor in Northwestern Indiana. This dataset contains 145×145 pixels, with 220 spectral bands in the wavelength range from 0.4 to 2.5 μm and is mainly composed of multiple agricultural fields. The spatial resolution is approximately 20 meters per pixel. Since removing bands covering the region of water absorption, 200 bands of the data were retained. After removing the background pixels, 10,249 pixels were reserved, which contain 16 classes representing the different land-cover types. Fig. 8 shows the false-color composite of the image and the corresponding ground truth.

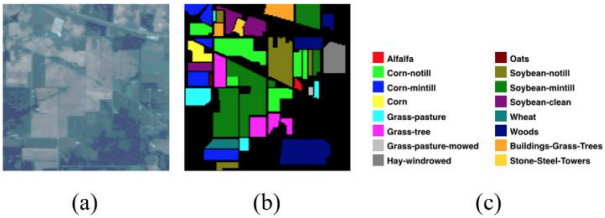

Fig. 8. The Indian Pines dataset. (a) Three-band false color composite. (b) Ground-truth map (c) Legend

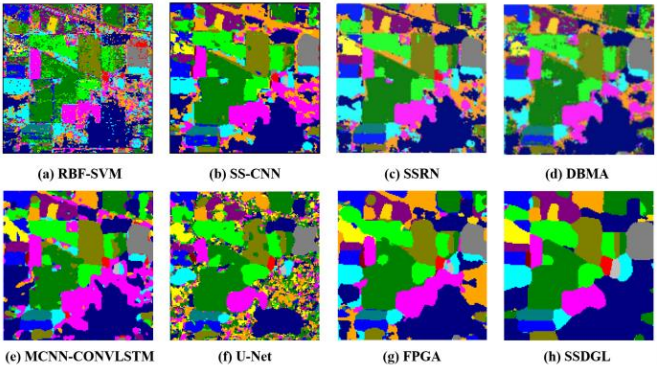

Fig. 9. Visualization of the classification maps for the Indian Pines dataset. (a) RBF-SVM. (b) SS-CNN. (c) SSRN. (d) DBMA. (e) MCNN-CONVLSTM. (f) U-Net (g) FPGA. (h) SSDGL.

Table I lists the number of training and testing data per class. The training samples were set to 5% of all labeled samples. If the training data of the class was less than 5, the mini-batch per class was set to 5. The training data were obtained by H-B sampling strategy and the remaining data were viewed as test data to evaluate the accuracy.

TABLE I
THE NUMBER OF TRAINING SAMPLES AND TEST SAMPLES FOR THE INDIAN PINES DATASET

| No. | Class. | Train. | Test. | Total. |
|-----|--------|--------|-------|--------|
| 1 | Alfalfa | 5 | 41 | 46 |
| 2 | Corn-notill | 72 | 1356 | 1428 |
| 3 | Corn-mintill | 42 | 788 | 830 |
| 4 | Corn | 12 | 225 | 237 |
| 5 | Grass-pasture | 25 | 458 | 483 |
| 6 | Grass-trees | 37 | 693 | 730 |
| 7 | Grass-pasture-mowed | 5 | 23 | 28 |
| 8 | Hay-windrowed | 24 | 454 | 478 |
| 9 | Oats | 5 | 15 | 20 |
| 10 | Soybean-notill | 49 | 923 | 972 |
| 11 | Soybean-mintill | 123 | 2332 | 2455 |
| 12 | Soybean-clean | 30 | 563 | 593 |
| 13 | Wheat | 11 | 194 | 205 |
| 14 | Woods | 64 | 1201 | 1265 |
| 15 | Buildings-Grass-Trees | 20 | 366 | 386 |
| 16 | Stone-Steel-Towers | 5 | 88 | 93 |
| | Total | 529 | 9720 | 10249 |

Fig. 9 (a)-(g) illustrates the classification results using RBF-SVM, SS-CNN, SSRN, DBMA, MCNN-CONVLSTM, U-Net and FPGA. It can be seen that the classification methods based on CNN had a better visual performance than SVM, and the image is smoother than SVM. This is because the convolutional neural network-based method considered the spatial features of adjacent pixels. The FPGA and SSDGL obtained the complete structure of land covers, and the category boundaries are closer to real images. This is because the FCN-based method makes full use of the global spatial context to extract the most discriminating spatial features. Compared with U-Net, the classification maps of FPGA and SSDGL show that these methods have better classification performance in the categories with similar spectral features, such as corn and soybean, and these methods have better generalization ability. GCL module played an important role in representation learning, and the correlation of the adjacent channels and long-range channels was simultaneously considered. It can be seen from Fig. 10 (a) that the large intraclass variations and the small interclass dissimilarity existed in the Indian Pines dataset. The proposed framework can obtain the most discriminative feature representations to reduce the intraclass distance and increase the interclass distance, as shown in Fig. 10 (b).

For a more detailed verification of the results, the overall accuracies (OA), average accuracy (AA), kappa coefficients, and per-class accuracies are presented in Table II for all classification methods (RBF-SVM, SS-CNN, SSRN, DBMA, MCNN-CONVLSTM, U-Net, and FPGA). The best accuracy is highlighted in bold for each row in the table.

TABLE II

THE CLASSIFICATION RESULTS OF RBF-SVM, SS-CNN, SSRN, DBMA, MCNN-CONVLSTM, U-NET, FPGA AND SSDGL ON THE INDIAN PINES DATASET WITH 5% LABELED SAMPLES.

| Class | CNN-based | | | | | FCN-based | | |
|-------|-----------|--------|--------|--------|------------------|-------|--------|----------|
|       | RBF-SVM   | SS-CNN | SSRN   | DBMA   | MCNN-CONVLSTM | U-Net | FPGA   | Proposed |
| 1     | 70.32     | 72.14  | 75.57  | 90.37  | 94.36            | 97.67 | 97.22  | **100.00** |
| 2     | 69.63     | 90.42  | 90.65  | 92.72  | 92.84            | 92.48 | 93.07  | **99.63** |
| 3     | 58.26     | 81.48  | 97.01  | 95.63  | 93.02            | 84.77 | 89.46  | **99.24** |
| 4     | 45.22     | 71.23  | 93.36  | 89.35  | 95.32            | 89.33 | 100.00 | **100.00** |
| 5     | 75.48     | 83.62  | 98.56  | 96.92  | 92.13            | 81.00 | 95.63  | **99.56** |
| 6     | 96.14     | 97.19  | 98.94  | 99.18  | 98.86            | 94.08 | 97.56  | **100.00** |
| 7     | 95.79     | 91.03  | 84.21  | 79.57  | 84.83            | 100.00| 100.00 | **100.00** |
| 8     | 87.72     | 92.34  | 98.36  | 99.11  | 98.63            | 98.90 | 100.00 | **100.00** |
| 9     | 75.03     | 96.39  | 97.61  | 97.91  | 92.47            | 78.95 | 100.00 | **100.00** |
| 10    | 66.25     | 81.75  | 81.03  | 92.08  | 94.76            | 89.49 | 96.64  | **99.68** |
| 11    | 77.62     | 87.39  | 93.02  | 95.15  | 96.28            | 97.81 | 96.74  | **99.36** |
| 12    | 67.28     | 83.03  | 95.72  | 90.71  | 94.12            | 86.50 | 91.65  | **99.11** |
| 13    | 96.93     | 97.42  | 99.81  | 99.81  | 96.95            | 98.97 | 100.00 | **100.00** |
| 14    | 95.07     | 95.31  | 95.79  | 97.11  | 98.79            | 98.58 | 99.91  | **100.00** |
| 15    | 35.48     | 74.04  | 92.25  | 88.13  | 92.83            | 92.08 | 99.72  | **100.00** |
| 16    | 97.61     | 94.61  | 96.57  | 97.05  | 87.32            | 93.18 | 100.00 | **100.00** |
| OA    | 75.31     | 89.82  | 92.21  | 94.43  | 94.78            | 93.20 | 96.18  | **99.63** |
| AA    | 71.12     | 83.73  | 93.03  | 93.81  | 93.37            | 92.11 | 97.33  | **99.79** |
| Kappa | 0.7173    | 0.8783 | 0.9115 | 0.9365 | 0.9437           | 0.9222| 0.9564 | **0.9958** |

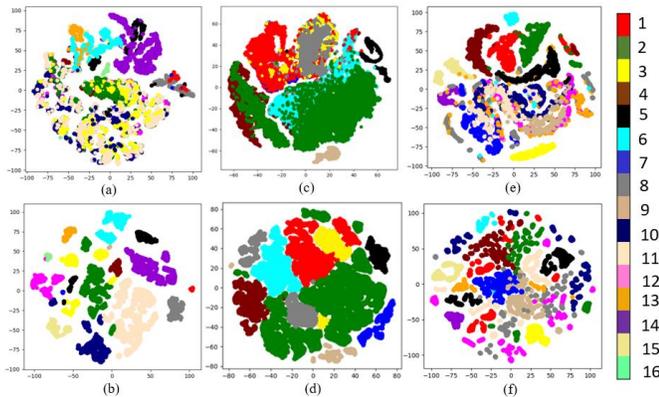

Fig. 10. Two-dimensional t-SNE visualization of features from the Indian Pines, Pavia University and Houston University datasets. Data distributions of the labeled samples in the original feature space (the first row) and the convolutional feature space (the second row). Different colors correspond to different classes.

As shown in Table II, FCN-based methods obtained better class accuracy, and the overall accuracy was above 90%. It is attributed to the global learning framework, which makes full use of the global spatial context information. Since agricultural land cover types have a large spectral difference between the green band and the near-infrared band, the proposed method can highlight the correlation between these bands. The classification accuracy on soybean using SSDGL achieves a 3%~7% improvement over FPGA and achieved a 6%~10% improvement on corn. Moreover, the hierarchically balanced sampling strategy played a key role in the class imbalance problem. The weighted softmax loss is employed to reduce the weight of easy-to-classify samples so that the model focused more on hard-to-classify samples during training. It can be seen in Table I that some number of training sample categories were less than 10, and some were more than 50. The categories with a small number of training samples were hard to classify, so the classification accuracy on corn, grass, and soybean were worse than other categories. The category weighting factor was added to the cross-entropy criterion function to balance the relative loss of well-to-classify samples and hard-to-classify samples. Hence, the proposed method achieved great classification performance on datasets with insufficient and imbalanced samples. It can be seen that the FCN-based method had higher accuracy on OA, AA, and kappa coefficient. Compared with FPGA, SSDGL achieved ~3% improvement in OA, AA and kappa coefficient. The novel sampling strategy and loss function can effectively solve the long-tail distribution problem of hyperspectral image datasets.

### C. Experiment 2: Pavia University Dataset

The Pavia University dataset was acquired by the Reflective Optics System Imaging Spectrometer (ROSIS) sensor over the University of Pavia in 2001. Since 12 bands covering the region of noise and water absorption were removed, 103 bands of the data were retained, and the spectral range was from 0.43 to 0.86 μm. This dataset has 610×340 pixels with a resolution of 1.3 meters per pixel. After removing the background pixels, 42,776 pixels were reserved, which contained nine classes representing the different land-cover types. Fig. 11 shows the false-color composite of the image and the corresponding ground truth. Table. III lists the number of training and test data for each category. We set the training sample number per class as 1% of labeled samples.

Fig. 12 (a)-(g) illustrates the classification results using RBF-SVM, SS-CNN, SSRN, DBMA, MCNN-CONVLSTM, U-Net and FPGA. It can be seen that the classification maps of RBF-SVM contain salt-pepper noise because this method only considers the spectral information and ignores the spatial correlation of the adjacent pixels. Therefore, the most discriminative features are difficult to extract, and the classification performance is limited. Due to the different materials of the roof, it is difficult to discriminate the building classes based on spatial context information, so the spectral features need to be emphasized. It can be seen from the classification maps that the building category labels had great differences with different methods, but the proposed method can obtain the complete shape of buildings and accurately discriminated the building materials. The GCL module was utilized to extract the interdependency between the channels according to the spectral information of the whole hyperspectral image. Because the global learning method can model the long-range dependency, the complete structure of the road can be obtained. The sample distribution of Pavia dataset is shown in Fig. 10 (c). There are many labeled samples in this dataset, but similar land cover types have difficulty distinguishing in the original HSI. The most discriminative features can be learned by the SSDGL, and the category boundary was accurately determined with generated feature maps.

### TABLE III
### The Number of Training Samples and Test Samples for the Pavia University Dataset

| No. | Class. | Train. | Test. | Total. |
|-----|--------|--------|-------|--------|
| 1 | Asphalt | 67 | 6564 | 6631 |
| 2 | Meadows | 187 | 18462 | 18649 |
| 3 | Gravel | 21 | 2078 | 2099 |
| 4 | Trees | 31 | 3033 | 3064 |
| 5 | Metal sheets | 14 | 1331 | 1345 |
| 6 | Bare Soil | 51 | 4978 | 5029 |
| 7 | Bitumen | 14 | 1316 | 1330 |
| 8 | Bricks | 37 | 3645 | 3682 |
| 9 | Shadows | 10 | 937 | 947 |
| | Total | 432 | 42344 | 42776 |

To evaluate the performance of these methods on this dataset from a quantitative perspective, the overall accuracies (OA), per-class accuracies (AA), and kappa coefficients are presented in Table IV for RBF-SVM, SS-CNN, SSRN, DBMA, MCNN-CONVLSTM, U-Net, and FPGA. The best accuracy is highlighted in bold for each row in the table. As shown in Table IV, the spatial resolution of this dataset is very high, so the spatial information is important for HSI to discriminate hard-to-classify categories. The classification accuracy varied greatly on gravel with different methods, but the SSDGL framework achieved the best classification accuracy and was ~10% than other methods. This is because the GCL module was introduced to SSDGL to extract the interdependency of channels according to continuous spectral sequence and global spatial context information.

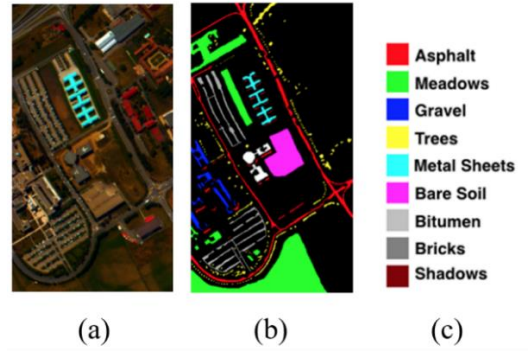

Fig. 11. The Pavia University dataset. (a) Three-band false color composite. (b) Ground-truth map (c) Legend

The classification accuracy of the FCN-based method was ~ 3% higher than that of the CNN-based method on bare soil and ~10% higher on bitumen. It can be concluded that the high spatial resolution remote sensing image facilitates spatial feature extraction and boosts the classification performance. The classification accuracy of brick reached 99.92% with the SSDGL framework and ~1.5% higher than that of the FPGA. This was attributed to the global joint attention mechanism (GJAM), which extracted fine-grained spatial features and attention areas. It can be observed that SSDGL had higher accuracy in OA, AA, and kappa coefficient. The proposed method was ~3% higher than that of the CNN-based methods and ~1% higher than that of the FPGA. Although the number of training samples was limited, the classification accuracy per class of the SSDGL reached 99% because the proposed method has strong feature learning ability, and it is good at classifying class imbalanced datasets.

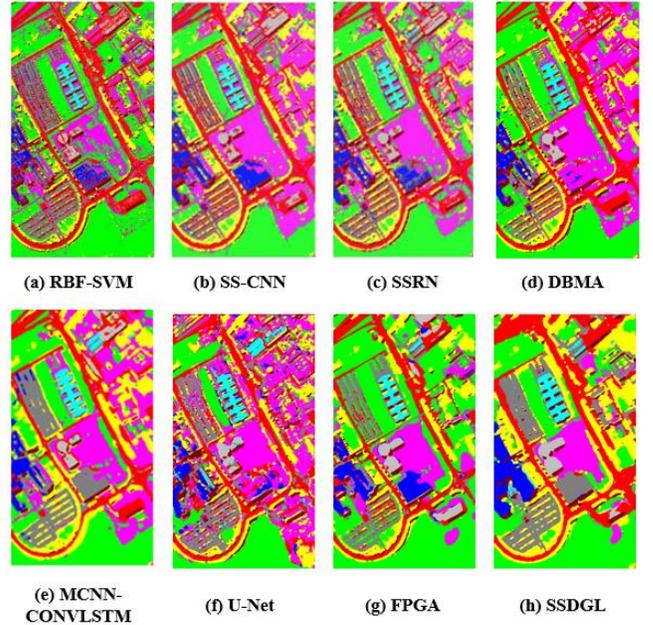

Fig. 12. Visualization of the classification maps for the Pavia University dataset. (a) RBF-SVM. (b) SS-CNN. (c) SSRN. (d) DBMA. (e) MCNN-CONVLSTM. (f) U-Net (g) FPGA. (h) SSDGL.

## D. Experiment 3: Houston University Dataset

The Houston University dataset is a public HSI dataset that was released in the 2013 IEEE GRSS Data Fusion Contest. The Houston University dataset is a more challenging hyperspectral dataset that was captured by the National Center for Airborne Laser Mapping (NCALM) over the Houston University campus and contains 15 complex land-cover classes with 349×1,905 pixels and 144 bands ranging from 0.36 to 1.05 µm. To further verify the validity of the proposed framework with limited training samples, the ten training samples per class was used to evaluate the performance of the proposed method. Fig. 13 shows the false-color composite of the image and the corresponding ground truth. Table. V lists the number of training and test data for each category. Since the number of training samples per class is ten, the strong performance of the proposed model can be presented in this case.

The classification performance of SSDGL is compared with seven state-of-the-art methods, which are summarized as follows: SVM-3DG [51], SS-CNN, SSRN, MugNet [9], AROC-DPNet [6], U-Net, and FPGA. SVM-3DG is a SVM-based method with 3D discrete wavelet transform and Markov random field. MugNet is a state-of-the-art deep learning method for small sample HSI classification. AROC-DPNet is a lightweight convolutional neural network with a deep clustering strategy. The weight decay rate was set to 0.001 and trained in 1,000 epochs.

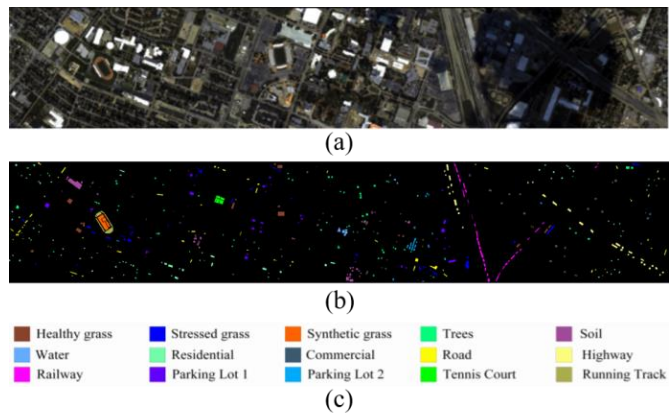

(a)

(b)

(c)

| | | | | | |
|---|---|---|---|---|---|
| ■ Healthy grass | ■ Stressed grass | ■ Synthetic grass | ■ Trees | ■ Soil | |
| ■ Water | ■ Residential | ■ Commercial | ■ Road | ■ Highway | |
| ■ Railway | ■ Parking Lot 1 | ■ Parking Lot 2 | ■ Tennis Court | ■ Running Track | |

Fig. 13. The Houston University dataset. (a) Three-band false color composite. (b) Ground-truth map (c) Legend

It can be seen in Table VI that the SSDGL framework achieve best classification performance than other popular methods. The overall accuracy was 10 ~ 20% higher than that of the CNN-based methods. Hence, the global learning framework had a remarkable breakthrough in the small sample classification of hyperspectral images. Compared with U-Net and FPGA, the proposed framework achieved a better classification accuracy on OA, AA, and Kappa. It can be concluded that an appropriate sampling strategy and loss function is necessary to address the insufficient sample problem. It can be seen from Fig. 10 (e) that the samples exist a small difference and some categories are difficult to distinguish in the original feature space. However, the proposed method has a strong feature learning ability and increase the gap between the different categories in the convolutional feature space. These categories can be easily distinguished on the generated spectral-spatial features.

TABLE V
THE NUMBER OF TRAINING SAMPLES AND TEST SAMPLES FOR THE HOUSTON UNIVERSITY DATASET

| No. | Class. | Train. | Test. | Total. |
|---|---|---|---|---|
| 1 | Healthy Grass | 10 | 1241 | 1251 |
| 2 | Stressed Grass | 10 | 1244 | 1254 |
| 3 | Synthetic Grass | 10 | 687 | 697 |
| 4 | Tree | 10 | 1234 | 1244 |
| 5 | Soil | 10 | 1242 | 1252 |
| 6 | Water | 10 | 315 | 325 |
| 7 | Residential | 10 | 1258 | 1268 |
| 8 | Commercial | 10 | 1234 | 1244 |
| 9 | Road | 10 | 1242 | 1252 |
| 10 | Highway | 10 | 1217 | 1227 |
| 11 | Railway | 10 | 1225 | 1235 |
| 12 | Parking Lot 1 | 10 | 1224 | 1234 |
| 13 | Parking Lot 2 | 10 | 459 | 469 |
| 14 | Tennis Court | 10 | 418 | 428 |
| 15 | Running Track | 10 | 650 | 660 |
| | Total | 150 | 14861 | 15011 |

The confusion matrix is shown in Fig. 14. It can be seen that the classification accuracy of each category was higher than 90%, except for the commercial class. The commercial class was easily misclassified as residential, road and parking lot because these categories have similar spectral-spatial features, and the fine-grained spatial difference struggled to extract with a small number of training samples. The SSDGL achieved great classification accuracy for some land-cover categories, so we did not use data augmentation to further improve the classification performance.

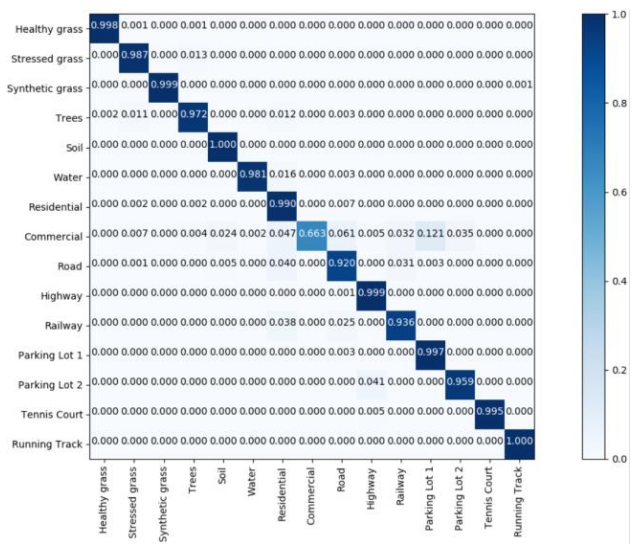

Fig. 14. Confusion matrix of SSDGL on the Houston University dataset.

TABLE IV

THE CLASSIFICATION RESULTS OF RBF-SVM, SS-CNN, SSRN, DBMA, MCNN-CONVLSTM, U-NET, FPGA AND SSDGL ON THE PAVIA UNIVERSITY DATASET WITH 1% LABELED SAMPLES

| Class | Patch-based | | | | | Patch-free | | |
|---|---|---|---|---|---|---|---|---|
| | RBF-SVM | SS-CNN | SSRN | DBMA | MCNN-CONVLSTM | U-Net | FPGA | Proposed |
| 1 | 85.82 | 92.21 | 97.41 | 96.26 | 96.53 | 94.06 | 97.83 | **100.00** |
| 2 | 96.02 | 90.27 | 99.10 | 98.31 | 99.26 | 98.67 | 99.95 | **100.00** |
| 3 | 65.46 | 79.82 | 88.61 | 89.16 | 87.15 | 78.30 | 91.28 | **100.00** |
| 4 | 81.24 | 92.67 | **99.81** | 97.53 | 93.68 | 96.01 | 95.85 | 99.67 |
| 5 | 99.23 | 99.41 | 100.00 | 99.72 | 97.26 | 100.00 | 100.00 | **100.00** |
| 6 | 67.69 | 86.86 | 95.51 | 97.98 | 96.41 | 99.80 | 99.76 | **100.00** |
| 7 | 53.86 | 79.85 | 92.16 | 85.51 | 88.74 | 79.56 | 99.73 | **100.00** |
| 8 | 86.28 | 92.83 | 89.03 | 80.70 | 93.81 | 99.58 | 98.05 | **99.92** |
| 9 | 99.92 | 93.93 | 99.96 | 91.12 | 93.79 | 99.25 | 97.86 | **100.00** |
| OA | 86.54 | 90.14 | 96.55 | 95.13 | 96.28 | 96.09 | 98.68 | **99.97** |
| AA | 73.52 | 83.52 | 95.73 | 93.48 | 94.69 | 93.47 | 97.82 | **99.95** |
| kappa | 0.8192 | 0.8768 | 0.9543 | 0.9353 | 0.9481 | 0.9480 | 0.9825 | **0.9996** |

TABLE VI

THE CLASSIFICATION RESULTS OF SVM-3DG, SS-CNN, SSRN, MUGNET, AROC-DPNET, U-NET, FPGA, AND SSDGL ON HOUSTON UNIVERSITY DATASET WITH TEN SAMPLES PER CLASS

| Class | SVM-3DG | SS-CNN | SSRN | MugNet | AROC-DPNet | U-Net | FPGA | Proposed |
|---|---|---|---|---|---|---|---|---|
| 1 | 73.88 | 96.74 | 74.37 | 87.21 | 98.82 | 84.53 | 97.90 | **99.84** |
| 2 | 68.65 | 93.14 | 78.41 | 86.66 | 90.75 | 48.79 | 87.54 | **98.71** |
| 3 | 66.43 | 98.52 | 73.26 | 97.58 | 99.59 | 99.85 | **100.00** | 99.85 |
| 4 | 78.83 | 79.59 | 78.53 | 80.41 | 93.31 | 74.72 | **98.22** | 97.16 |
| 5 | 86.37 | 93.64 | 92.56 | 97.87 | 99.65 | 87.34 | 99.92 | **100.00** |
| 6 | 66.82 | 95.53 | 76.94 | 88.25 | 96.78 | 97.78 | 93.97 | **98.10** |
| 7 | 61.43 | 77.46 | 54.27 | 69.63 | 91.23 | 80.68 | 84.34 | **98.97** |
| 8 | 44.72 | 61.28 | **92.89** | 62.82 | 88.92 | 44.25 | 66.12 | 66.29 |
| 9 | 48.53 | 88.79 | 68.58 | 79.91 | 87.85 | 37.04 | 91.06 | **92.03** |
| 10 | 67.18 | 86.73 | 65.11 | 92.18 | 90.72 | 78.55 | 96.63 | **99.92** |
| 11 | 58.42 | 36.41 | 82.07 | 83.15 | 65.03 | 60.24 | 68.82 | **93.63** |
| 12 | 52.71 | 75.06 | 74.63 | 78.75 | 95.79 | 73.83 | 90.60 | **99.67** |
| 13 | 81.16 | 28.48 | 73.77 | 89.81 | 47.16 | 92.16 | **96.29** | 95.86 |
| 14 | 84.93 | 91.34 | 74.83 | 98.17 | 76.29 | 100.00 | 100.00 | **100.00** |
| 15 | 69.53 | 96.08 | 89.12 | 97.20 | 94.83 | 72.77 | 100.00 | **100.00** |
| OA | 64.51 | 78.63 | 74.51 | 84.23 | 88.93 | 71.11 | 89.89 | **95.36** |
| AA | 67.36 | 78.82 | 77.18 | 85.81 | 87.19 | 75.50 | 91.43 | **96.00** |
| Kappa | 0.6287 | 0.7762 | 0.7235 | 0.8413 | 0.8781 | 0.6883 | 0.8908 | **0.9499** |

## V. DISCUSSION

To better understand the effectiveness of each component in the spectral-spatial dependent global learning (SSDGL) framework, we conducted extensive analysis experiments of each module. All component analysis experiments were performed on the Indian Pine dataset, which has a low spatial resolution and the sample data is imbalanced. The baseline method is shown in Table. VII (a), which is an encoder-decoder architecture (SegNet) trained by the global stochastic stratified (GS²) sampling strategy of FPGA.

### A. Discussion on the GCL module and GJAM module

The global convolutional long short-term memory (GCL) is described in Section III. Table. VII (b) presents the classification performance of the baseline method with the GCL module. The GCL module is added to SSDGL, OA increased from 58.39% to 95.69%. The hyperparameter $\alpha$ is introduced into the GCL module to determine the number of time steps. Where the value of $\alpha$ is from 4 to 12, and the interval is 2. When $\alpha$ is set to 8, the best classification accuracy can be obtained by the extracted spectral-dependent features.

Table. VII (c) presents the effectiveness of the global joint attention module (GJAM). The addition of GJAM modules to SSDGL results in an OA improvement from 95.69% to 96.97%. This module reweight the feature maps and extracts attention areas to boost the classification performance. The main hyperparameter is a compression factor in the global spectral attention mechanism. Due to the limitation of the space, we do not show the results of various parameters.

### B. Discussion on the H-B sampling strategy and weighted softmax loss

Table VII (d) presents the results of the SSDGL with the H-B sampling strategy and weighted softmax loss (HB-WL). The OA is improved from 96.77 to 99.63 and AA from 97.65 to 99.79. The H-B sampling strategy is used to balance the number of each class in the hierarchical training samples. The category probability is recalculated by the weighted softmax loss.

The hyperparameter $\beta$ is introduced to the H-B sampling strategy to control the mini-batch per class of hierarchical training samples. The value of $\beta$ ranges from 5 to 40 and the interval is 5. The proposed SSDGL has the best classification performance when $\beta$ is set to 10. When $\beta$ is set to a value larger than 20, the classification accuracy of SSDGL decreases gradually, and the average accuracy is limited. It can be found that the value of $\beta$ has little impact on the classification performance because the weighted softmax loss recalculates the category probability, but the H-B sampling strategy still plays a key role in addressing insufficient and imbalanced sample problems.

TABLE VII
HSI CLASSIFICATION RESULTS EVALUATED ON THE INDIAN PINES DATASET WITH 5% LABELED SAMPLES. THE ENCODER-DECODER BASELINE, THE GCL, GJAM, AND HB-WL ARE ADDED IN SSDGL FOR THE MODULE ANALYSIS

| Method | GCL | GJAM | HB-WL | OA | AA | Kappa |
|---|---|---|---|---|---|---|
| (a) Baseline | - | - | - | 58.39 | 59.26 | 0.5365 |
| (b) SSDGL w/o GJAM and HB-WL | √ | - | - | 95.69 | 96.14 | 0.9532 |
| (c) SSDGL w/o HB-WL | √ | √ | - | 96.77 | 97.65 | 0.9649 |
| (d) SSDGL | √ | √ | √ | 99.63 | 99.79 | 0.9958 |

### C. Discussion on the running time

Table VIII lists the training and testing time of the five methods on the Indian pines (IP), Pavia University (PU), and Houston University (HU) datasets. Although SSRN designed with the deep CNN, the training speed of SSRN was 2 times faster than that of SS-CNN. The semisupervised classification methods may achieve better performance than the supervised classification methods when the training samples are small, but they consume considerable computing memory and time.

TABLE VIII
TRAINING AND TESTING TIME OF DIFFERENT MODELS ON THREE HSI DATA SETS

| | | IN | PU | HU |
|---|---|---|---|---|
| SS-CNN | Train.(m) | 25.7 | 24.8 | 37.3 |
| | Test.(s) | 19.6 | 35.4 | 45.2 |
| SSRN | Train.(m) | 18.5 | 16.8 | 21.7 |
| | Test.(s) | 15.2 | 29.6 | 36.5 |
| U-Net | Train.(m) | 4.7 | 4.2 | 15.2 |
| | Test.(s) | 0.23 | 0.27 | 0.41 |
| FPGA | Train.(m) | 4.2 | 3.9 | 14.6 |
| | Test.(s) | 0.16 | 0.19 | 0.34 |
| SSDGL | Train.(m) | 9.8 | 8.9 | 27.5 |
| | Test.(s) | 0.31 | 0.36 | 0.63 |

It can be seen that FCN-based methods reduce memory and time consumption. We set the training iterations of the IP and PU datasets in 600 epochs and set 1,000 epochs on the HU datasets to make the model converge. The training speed of the global learning methods faster than patch-based deep learning methods and the testing speed was 100 times faster than that of patch-based methods. Moreover, the test data of global learning methods is a whole image rather than a patch of each test pixel, so the model inference time is greatly decreased because the global learning method reduces the redundant calculations of the overlapping areas between the adjacent pixel patches. The training times of SSDGL are 2~3 times longer than those of the U-Net and FPGA, but the classification accuracy achieved a significant improvement. Hence, the proposed framework is valuable and has good application prospects.

### VI. CONCLUSION

In this paper, a spectral-spatial dependent global learning (SSDGL) framework is proposed to improve the classification performance of hyperspectral images (HSI) with insufficient and imbalanced samples. In the SSDGL framework, the hierarchically balanced (H-B) sampling strategy is proposed to divide the training data into some hierarchical training samples. The weighted softmax with cross entropy loss is used to recalculate the category probability according to the number of labeled samples per class. The input data of SSDGL is the whole image, and it does not require dividing the HSI dataset into pixel patches. To extract the interdependence between spectral features, the global convolutional long short-term memory (GCL) is added to the SSDGL. The global joint attention mechanism (GJAM) is used to estimate the importance of different spectral-spatial features. It can be seen from the experimental results that SSDGL achieves better classification performance than other state-of-the-art methods on HSI datasets, especially when the training samples are insufficient or imbalanced. Compared with the CNN-based methods, the global learning methods greatly reduce the training and inference time, thereby broadening its application prospects in HSI classification.